\documentclass[sigconf]{acmart}

\renewcommand\footnotetextcopyrightpermission[1]{} 
\usepackage{graphicx}
\usepackage{color,soul}
\usepackage{subcaption}
\usepackage{algorithm}
\usepackage{algorithmic}
\usepackage{multirow}
\usepackage{graphicx}
\usepackage[normalem]{ulem}
\useunder{\uline}{\ul}{}
\usepackage{enumitem}
\usepackage{hyperref}

\makeatletter
\renewcommand{\fnum@figure}{Fig.\thefigure}
\makeatother

\AtBeginDocument{%
  \providecommand\BibTeX{{%
    \normalfont B\kern-0.5em{\scshape i\kern-0.25em b}\kern-0.8em\TeX}}}

\setcopyright{acmcopyright}
\copyrightyear{2020}
\acmYear{2020}

\acmConference[Mobiquitous '20]{ 17th EAI International Conference on Mobile and Ubiquitous Systems: Computing, Networking and Services}{November 09--11, 2020}{NA, CyberSpace}
\acmBooktitle{17th EAI International Conference on Mobile and Ubiquitous Systems: Computing, Networking and Services,
  November 09--11, 2020, NA, Cyberspace}

\setcopyright{none}
\begin{document}

\title{Formation-based Selection of Drone Swarm Services}


\author{Balsam Alkouz}
\affiliation{%
  \institution{The University of Sydney}
  \country{Australia}}
\email{balsam.alkouz@sydney.edu.au}

\author{Athman Bouguettaya}
\affiliation{%
  \institution{The University of Sydney}
  \country{Australia}}
\email{athman.bouguettaya@sydney.edu.au}

\renewcommand{\shortauthors}{Alkouz and Bouguettaya}

\begin{abstract}
 Swarm of drones are increasingly being asked to carry out missions that can't be completed by one drone. Particularly, in delivery, issues arise due to the swarm's limited flight endurance. Hence, we propose a novel formation-guided framework for selecting Swarm-based Drone-as-a-Service (SDaaS) for delivery. A detailed study is carried out to highlight the effect of swarm formations on energy consumption. Two SDaaS selection approaches, i.e. Fixed and Adaptive, are designed considering the different formation decisions a swarm can take. The proposed framework considers extrinsic constraints including wind speed and direction. We propose SDaaS selection algorithms for each approach. Experimental results prove the efficiency of the proposed algorithms.
\end{abstract}



\maketitle

\section{Introduction}

Drones are one of the important IoT devices whose power is leveraged while developing and operating a smart city \cite{khan2018drones}. Drones or Unmanned aerial vehicles (UAVs) are aircrafts that can be navigated without a human pilot on board. The UAV technology has been evolving continuously over the past decade \cite{khuwaja2018survey}. This development and the low cost of the technology allowed hobbyists, entrepreneurs, and businesses to find various ways to leverage the drones technology. Multiple industries see an opportunity to cut costs and save time using drones technologies. \looseness=-1

There is an increasing need to use \emph{drone swarms} in a large number of applications. \emph{We define a swarm as a set of drones that act as a single entity to achieve the same goal.} Drone swarms are used in, but not limited to, military, sky shows, airborne communication networks, and delivery. There are several aspects that a drones' swarm should consider to deliver its potential. This includes collaboration, communication, and formation \cite{akram2017security}\cite{milani2016impact}. \looseness=-1

Drones \emph{swarm formation} is key to achieving a safe and efficient flights. We identify five swarm formations that drones in a swarm might take: Column, Front, Echelon, Vee, and Diamond. Flight formations are also studied in biology and military \cite{lissaman1970formation}\cite{haissig2004military}. In military for instance, different formations allow for a better view and more protection for individual planes \cite{haissig2004military}. In biology, the studies focus on the many species of birds that travel together in different formations \cite{lissaman1970formation}. The different formations are found to be taken for different purposes. For example, a Vee formation of geese help the birds to use less energy as they fly. This is because of their position in the swarm that lifts them up using the upwash forces generated by the front birds \cite{cutts1994energy}.\looseness=-1

This work focuses on using drone swarms \emph{for delivery purposes}. Companies including Amazon\footnote{Amazon Prime Air. https://www.amazon.com/Amazon-Prime-Air/}, Google\footnote{Google Wing - Canberra. https://wing.com/australia/canberra/}, and Uber\footnote{ Uber Eats Delivery. https://techcrunch.com/2019/06/12/uber-will-start-testing-eats-drone-delivery/} started advanced trials in drone delivery. Drone delivery is characterized by its fast speed and low cost \cite{shahzaad2019constraint}. The use of drone swarms in delivery adds to these benefits. For example, a single drone delivery has payload limitations, which can be overcome by the use of multiple drones. This is especially needed when multiple packages need to be delivered to the same destination at the same time e.g., emergencies. The drones in a swarm could also be made to deliver to \emph{farther destinations}. In this case, the payload that a single drone could carry may be divided among multiple drones to reduce the strain on the rotors, which directly affect the power consumption. This in result will increase the flight range of the swarm.\looseness=-1

We use the service paradigm as a way to model the delivery of services. We adopt a model for Swarm-based Drone-as-a-Service (SDaaS) where the authors abstract \emph{a swarm travelling in a line segment in a skyway network as a service} \cite{2020arXiv200506952A}. The service paradigm matches with the purpose of swarm delivery, i.e. the functional aspect. The function of a drone swarm delivery is to deliver multiple packages from a source to a destination. The non-functional (QoS) aspects of swarm delivery services include the delivery time, power consumption, and cost. Given a skyway network, \emph{our aim is to select the best skyway segments for delivery}. The best set is the set that optimizes the QoS properties. In this paper, we focus on optimizing the \emph{energy consumption}. Service selection results in a set of services, i.e. skyway segments, that are ready to be composed from a source to a destination. The extension of this work will be on the optimal composition of the selected services proposed here.\looseness=-1

Different from ground transportation such as trucks, drones are largely affected by wind \cite{takegami2019low}. In addition, they are highly constrained by battery capacities. In this paper, \emph{we explore the effect of different flight formations of drone swarms on the energy consumption}. We study the effect on the overall swarm energy consumption and on the individual drones. Our goal is to select the best skyway segments through flight formations. The formation of the swarm will guide the selection process. We propose an \emph{adaptive swarm formation} that considers the different \textit{wind speeds and wind directions.} The formations are considered adaptive as the swarm reforms itself when needed at every skyway segment. \looseness=-1

We summarize the key contributions of this research as follows:
\begin{itemize}
    \item A formation-guided Swarm-based Drone-as-a-Service (SDaaS) selection model.
    \item A detailed study on forces affecting swarms energy consumption under different wind conditions and formations.
    \item SDaaS selection algorithms for different types of SDaaS selections: Fixed and Adaptive. These types are derived from different types of formation decisions a swarm can make.
\end{itemize}

\subsection{Motivating Scenario}

\begin{figure}[ht]
\centering
\includegraphics[width=\linewidth]{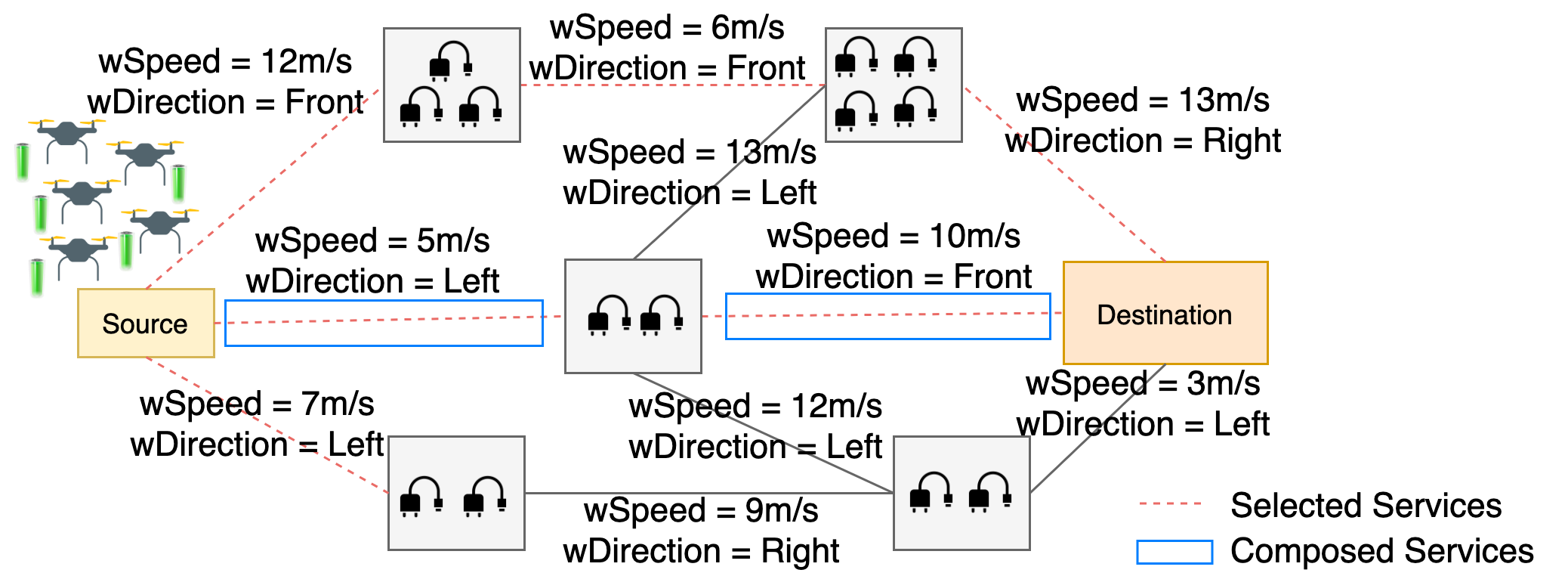}
\caption{Skyway network for swarm-based drone delivery}
\label{motivitaing}
\end{figure}

Let us assume a hospital uses drone delivery services to get medical equipment delivered on a regular basis. These pieces of equipment must be delivered together \emph{at the same time} on request. In this case, a swarm of drones would be needed to deliver multiple packages to the hospital by a certain deadline. In this context, a swarm of drones would use a skyway network whose vertices would consist of building rooftops with charging stations (see Fig. \ref{motivitaing}). The main challenge is to determine the best skyway path to deliver all packages within the expected delivery time. In addition, \emph{extrinsic constraints} would have to be addressed during medical supplies delivery. These constraints include the limited number of charging stations at intermediate nodes. The drones need to recharge due to their limited flight ranges. Another extrinsic constraint is the existence of environmental uncertainties such as wind. Therefore, \emph{we reformulate the problem of delivery as finding the best service paths in the skyway given these constraints}.

The first task is to select the best set of drones that can meet the deadline while carrying the equipment. The second task is to select the best formation. The swarm will rely on its formation to have an efficient delivery. In this paper we assume the drones set is selected and we focus on the second task.  We investigate \emph{the effect of different drones formations in a swarm under different wind conditions}. Drones energy consumption is affected heavily by wind conditions \cite{takegami2019low}. We propose to implement \emph{adaptive swarm formations} with the goal of reducing the energy consumption under different wind conditions. The swarm needs to \emph{minimize the energy consumption} to reduce the stops for battery recharges. This in return, will reduce the overall delivery time by reducing the \emph{charging and waiting times} caused by sequential recharges at each node. Each segment in a skyway network may have a different wind speed and wind direction. Therefore, \emph{we aim to \textbf{select} the best possible swarm formation for every wind condition}. This will guide the $SDaaS$ services (skyway segments) selection process.

\section{Related Work}

Drone swarms are found used in multiple applications. Some of these applications include target search \cite{avvenuti2018detection}, environmental monitoring \cite{chmaj2015distributed}, and delivery \cite{2020arXiv200506952A}. In target search, different targets applications has been implemented., e.g., toxic cloud, parasites, or humans \cite{avvenuti2018detection}. Swarms are being used in environmental monitoring applications including wildfire monitoring to help in fire-fighting operations in hazardous situations \cite{chmaj2015distributed}. A new emergent application for drone swarms is their use for delivery services. This has the benefit of carrying heavier payloads and the ability to travel longer compared to a single drone delivery \cite{2020arXiv200506952A}. 


Drone swarms can be classified by their different characteristics. These characteristics include behavior, movement, and formations. In terms of behavior, a swarm may be classified as static or dynamic \cite{akram2017security}. A static swarm is a swarm whose members are determined at the source point and no changes occur during the trip. A dynamic swarm is a swarm that allows the addition of new members and the leave of existing ones. In terms of movement, some of the most implemented movements include Stigmergy, Flocking, and Random movements \cite{cimino2015combining}. A swarm of drones may also be classified in terms of formations. \emph{We adopt the concept of formation flights from different domains and map it to drone swarms}. \looseness=-1


The study of flight formations has been widely explored in the literature. In military, formation flying is the flight were two or more aircrafts are under the command of a flight leader. Formation flying is used in military for defense purposes, better views, and concentration of firepower \cite{haissig2004military}. Formation flights are also seen in migrating birds. The Vee formation that migrating birds take enhances the overall aerodynamic efficiency by reducing the drag and thereby increasing the flight range \cite{cutts1994energy}. The study of flight formations in UAVs is still in its infancy. One application introduces the use of different geometrical flight formations in 3D scene reconstructions \cite{milani2016impact}. For energy conservation purposes, using a Vee formation in a fixed wing UAV swarm has proven to behave similar to  birds \cite{mirzaeinia2019energy}.  However, to the best of our knowledge the study of other types of formations for energy conservation purposes has not been explored yet. In addition, \emph{we could not find a work that focuses on the flight formations of quacopters specifically}. Quadcopters are the one of the most popular types of UAVs that are currently being used in various applications including delivery. \looseness=-1



There have already been an increasing interest in using UAVs in commercial purposes, particularly in delivery. However, most of these attempts use a single drone for the delivery queries. Few works addressed the delivery using a swarm of drones. Nevertheless, these works consider coordinating a set of single independent travelling drones that deliver different requests as a swarm \cite{san2016delivery}. In contrast, we define a swarm as a set of drones that act as a single entity and travel together to achieve their goal. \looseness=-1


The study of energy consumption in drones, specifically quadcopters has been carried in multiple works.  A power model was proposed in \cite{maekawa2017power} to measure power consumption in horizontal flights. The authors concluded that in a horizontal flight the weight load directly affects the power consumption. However, the flight speed does not have a big impact on the power consumption. Another work introduces a low energy routing model for drone delivery under windy conditions \cite{takegami2019low}. They proposed a routing algorithm to meet delivery deadlines. However, the work considered static wind conditions which is not realistic in real world scenarios. \emph{To the best of our knowledge most of the drone energy models in the literature deal with a single drone.} Our focus is to see the effect of drone swarm formations on energy consumption. \looseness=-1

The service paradigm is leveraged to model different types of travel services \cite{neiat2017crowdsourced}. A Drone-as-a-Service model using a single drone was also proposed. The objective of their work is to compose a set of drone services from a source to a destination with minimum delivery time. The authors addressed the constraints that surround a drone delivery environment \cite{shahzaad2019constraint}.  To the best of our knowledge, there is no previous work done on selecting swarm-based delivery services based on swarm formations. \emph{This paper is hence, the first attempt to model energy-constrained delivery environment for swarm-based drone deliveries under different wind conditions and formations.} \looseness=-1

\section{Swarm-based Drone-as-a-Service (SDaaS) Selection Model}

Service selection is the process of identifying services from all available services that match the functional and non-functional requirements of a consumer. An $SDaaS$ is defined as a swarm travelling in a skyway segment in a skyway network. The functional requirement is the successful travel in a skyway segment. The non-functional requirement is to travel with minimum energy consumption. 

\emph{Service selection is a step usually done before service composition to filter out services that are not optimal for the composition purposes}. In our scenario, service selection results in a set of services $Sel_{SDaaS}$, i.e. skyway segments, that are ready to be composed. The selected services are the set of services where all the drones in the swarm can travel successfully in a skyway segment without running out of battery power during the travel (Fig.\ref{motivitaing}). In the service composition step, the best set of services $Comp_{SDaaS}$ from the selected services are composed from the source to the destination (Fig.\ref{motivitaing}). \emph{The focus of this work is the selection of $SDaaS$.} Composition of $SDaaS$ will be the extension of this work.

\emph{We abstract each swarm travelling on a skyway between two nodes as an $SDaaS$ service} (see Fig. \ref{motivitaing}). In this paper, we consider the environment to be deterministic. \textbf{We formally define a Swarm-based Drone-as-a-a-Service (SDaaS) as a tuple of $<SDaaS\_id, S, F>$, where}
\begin{itemize}
    \item $SDaaS\_id$ is a unique identifier.
    \item $S$ is the swarm travelling in $SDaaS$. $S$ consists of $D$ which is the set of drones forming $S$, a tuple of $D$ is presented as $<d_1,d_2,..,d_n>$. $S$ also contains the properties including battery levels of every $d$ in $D$ $<b_1,b_2, ..,b_n>$, the payloads every $d$ in $D$ is carrying $<p_1,p_2,..,p_n>$, and the current node $N$ the swarm $S$ is at.\looseness=-1
    \item $F$ is the delivery function of a swarm on a skyway segment between two nodes, $A$ and $B$. $F$ consists of the travel time $tt$,  charging time $ct$, and waiting time $wt$ when recharging pads are not enough to serve all $D$ in node $B$.\looseness=-1
\end{itemize}

In this work, we focus on swarms whose members stay fixed throughout the delivery journey without splitting midway \cite{akram2017security}. The effect of flight of formations is more visible as the whole set travels together as a single entity. In this paper, we identify two types of swarms based on the formations decision: Preset swarms, and Flexible swarms. We explain each type below.

\subsection{Types of swarms}
\label{types}


\begin{figure*}[t!]
\centering

\subfloat[Preset Swarm]{\includegraphics[width=0.46\linewidth]{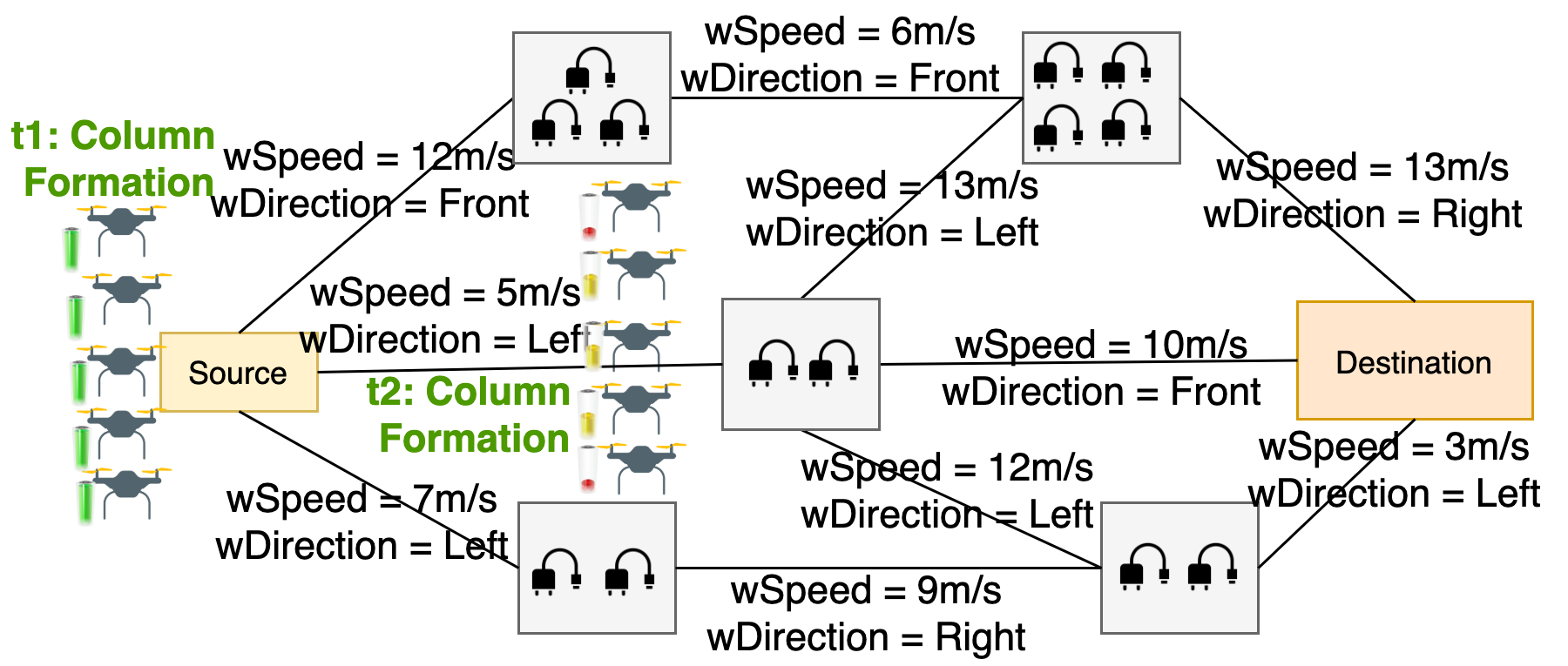}\label{fig:preset}
}\quad
\subfloat[Flexible Swarm]{\includegraphics[width=0.46\linewidth]{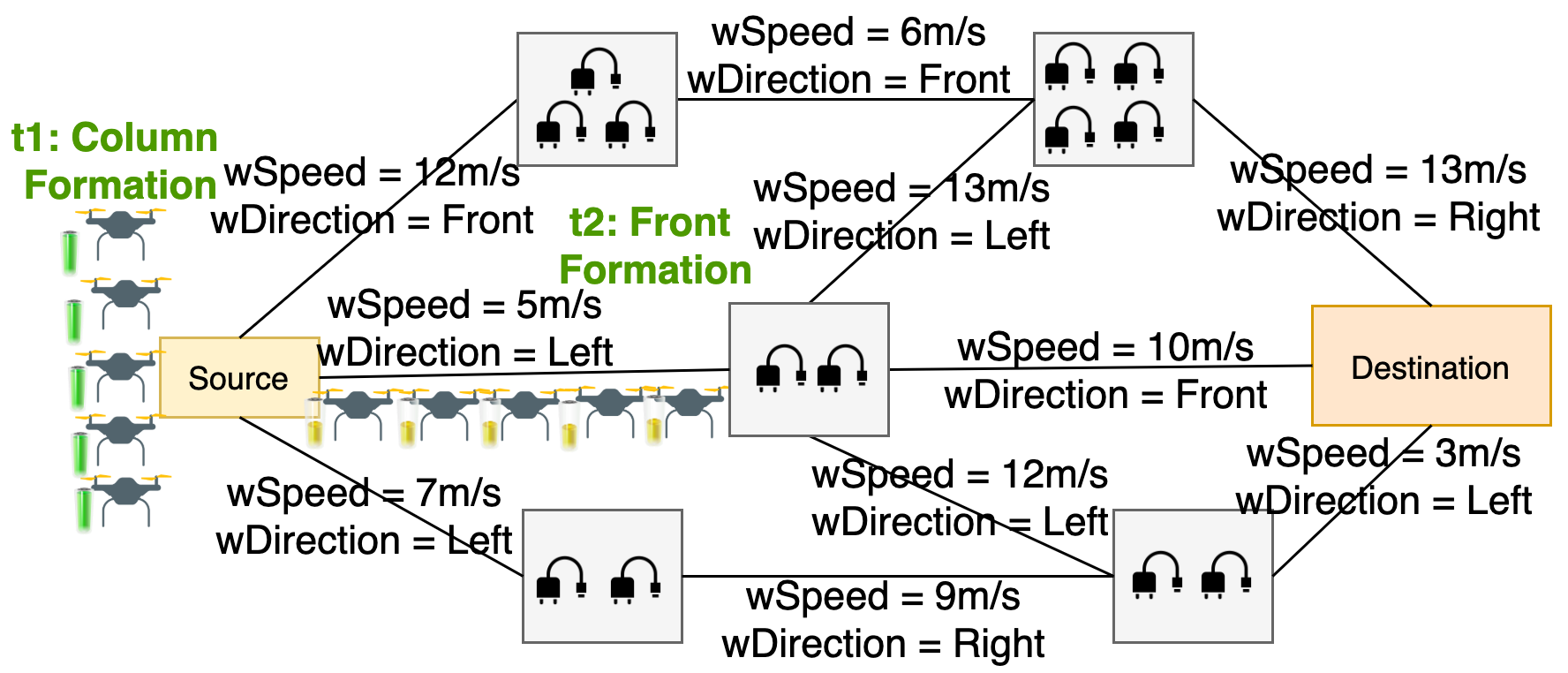}\label{fig:flexible}
}
\caption{Preset and Flexible swarm-based services snapshots at different times}
\label{fig:test}
\end{figure*}

\begin{itemize}[leftmargin=*]
    \item \textbf{Preset swarms:}
    A preset swarm is a swarm whose formation is \emph{determined before the selection and composition processes}. The decision of the formation is made before the travel is initiated and continues throughout the trip without changing (Fig.\ref{fig:preset}). In this case, selecting the optimal formation relies on the average wind conditions of all $SDaaS$ services. Computations regarding the wind effect on the power consumption are based on the pre-decided swarm formation. The selection eliminates $SDaaS$ where the swarm cannot travel in its pre-decided formation. Services in which the drones battery consumption is not sufficient to continue the travel safely are discarded. The selection process in this type is straight forward and does not require heavy computations during the service selection process. However, it comes with the cost of higher energy consumption. This is because the selection depends on the average wind condition and not the specific conditions in every skyway segment. We refer to this type of selection that deals with preset swarms as \emph{Fixed Selection}.\looseness=-1
   
    \item \textbf{Flexible swarms:}
    A flexible swarm is a swarm whose formation decisions are made throughout the trip. For every skyway segment, the swarm decides the best formation to take based on the wind condition of the segment. Hence, the swarm does not travel from the source to the destination in one formation. The formation changes as the wind condition changes (Fig.\ref{fig:flexible}). As mentioned earlier, for every wind condition there is an optimal flight formation that reduces the overall energy consumption of a swarm. Computations regarding the wind effect on the power consumption will be different at each segment based on the decided formation. The selection process eliminates $SDaaS$ services that the swarm cannot travel in regardless of the formation it takes. Services in which the drones battery consumption is not sufficient to continue the travel safely are discarded. This type of swarm overcomes the problem of higher energy consumption due to averaged conditions. We refer to this type of selection that deals with flexible swarms as \emph{Adaptive Selection}.\looseness=-1
    
\end{itemize}

\section{SDaaS Selection Framework}

In this section, we introduce the different forces affecting a swarm through different formations. We specifically study the upwash/downwash forces and the drag forces on a swarm. Then, we define the two types of selection algorithms: Fixed and Adaptive.


\subsection{Forces on a Drones Swarm} 
Computational fluid dynamics is a branch of fluid mechanics that uses numerical analysis and data structures to analyze and solve problems that involve fluid flows \cite{anderson1995computational}. In our case, we consider the wind as the fluid that we analyse its effect on drone swarms. We vary the conditions of wind including speed and direction. We adopt Ansys Discovery Live \footnote{Ansys Discovery Live. https://www.ansys.com/products/3d-design/ansys-discovery-live} to simulate the different formations under different conditions. Ansys Dicovery Live allows us to compute the drag force on all the faces of the drone model. This drag value is used to compute the power consumption using the following equation:

\begin{equation}
P= F_{D} . v
\label{equation}
\end{equation}

where $P$ is the power in Watts, $F_D$ is the drag force in Newtons, and $v$ is the velocity of the drone in meters/sec.

\begin{figure}[ht]
\centering
\includegraphics[width=\linewidth]{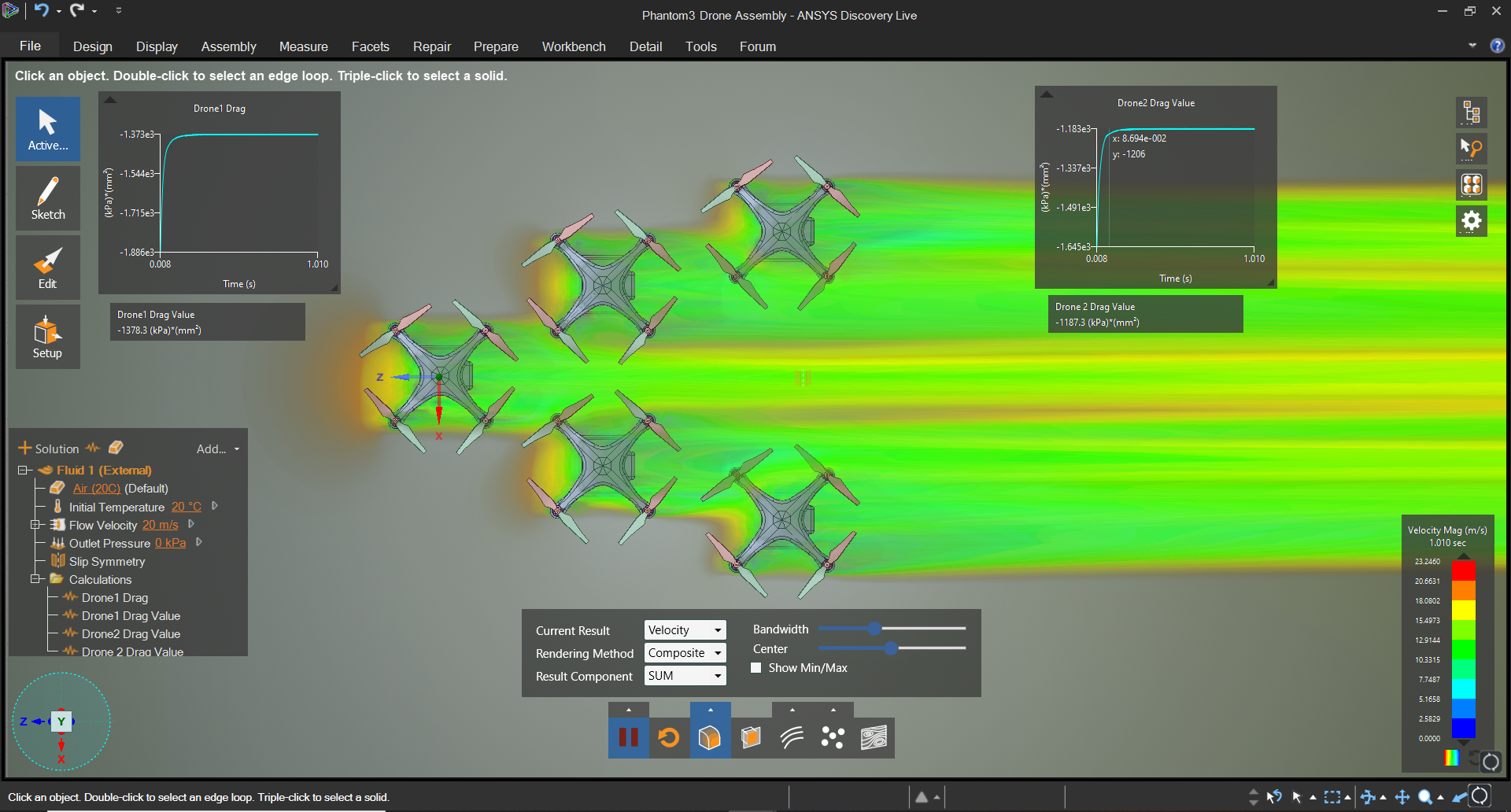}
\caption{Vee Formation in Ansys Discovery Live}
\label{ansysv}
\end{figure}

In this paper, we focus on modeling the energy consumption of different flight formations using quadcoptors. Although there are many types of drones, including fixed-wing and single-rotor helicopters, a quacopter is the most popular design for small UAVs. A \emph{quacopter}, or a quadrotor, is a multirotor helicopter lifted and propelled by four rotors. Big companies like DJI, Parrot, and 3DR are churning out quadcopters to take on more roles. One of the main reasons for the popularity of quadcopters is its mechanical simplicity to take-off and hover compared to other drones. In addition, the cost of their production is lesser compared to other drone types \cite{pounds2006modelling}. In this work, we use the DJI Phantom 3 Model for all our experiments.


We use the Beaufort wind force scale to categorize the wind speeds. This scale relates wind speed to observed conditions at sea or on land \cite{huler2007defining}. Table \ref{tab:my-table} shows the different wind speeds using Beaufort scale. We only consider Beaufort numbers 0 to 6 as the most commercial quadcopters are safe to fly under 13.8 m/s wind speed. We also consider three directions respective to the swarm origin including Front, Right, and Left. In the following subsections, we show the effect of the upwash and downwash forces generated by front drones in the swarm. We then test the different swarm formations under different wind conditions. This will help us select the best formation for every wind condition to be used in the $SDaaS$ service selection phase.

\begin{table}[ht]
\caption{Beaufort Wind Force Scale}
\label{tab:my-table}
\centering
\resizebox{\linewidth}{!}{%
\begin{tabular}{|l|l|l|l|}
\hline
\begin{tabular}[c]{@{}l@{}}Beaufort \\ Number\end{tabular} & Description     & \begin{tabular}[c]{@{}l@{}}Wind \\ Speed (m/s)\end{tabular} & Appearance                        \\ \hline
0                                                          & Calm            & \textless 0.5                                               & Still                             \\
1                                                          & Light air       & 0.5 - 1.5                                                   & Still                             \\
2                                                          & Light breeze    & 1.6 - 3.3                                                   & Leaves rustle                     \\
3                                                          & Gentle breeze   & 3.4 - 5.5                                                   & Leaves and small twigs move       \\
4                                                          & Moderate breeze & 5.5 - 7.9                                                   & Small branches move               \\
5                                                          & Fresh breeze    & 8 - 10.7                                                    & Small trees in leaf begin to sway \\
6                                                          & Strong breeze   & 10.8 - 13.8                                                 & Larger branches shake            \\ \hline        
\end{tabular}%
}
\end{table}

\subsubsection{Downwash and Upwash Forces.}

\begin{figure}
\centering
\includegraphics[width=0.7\linewidth]{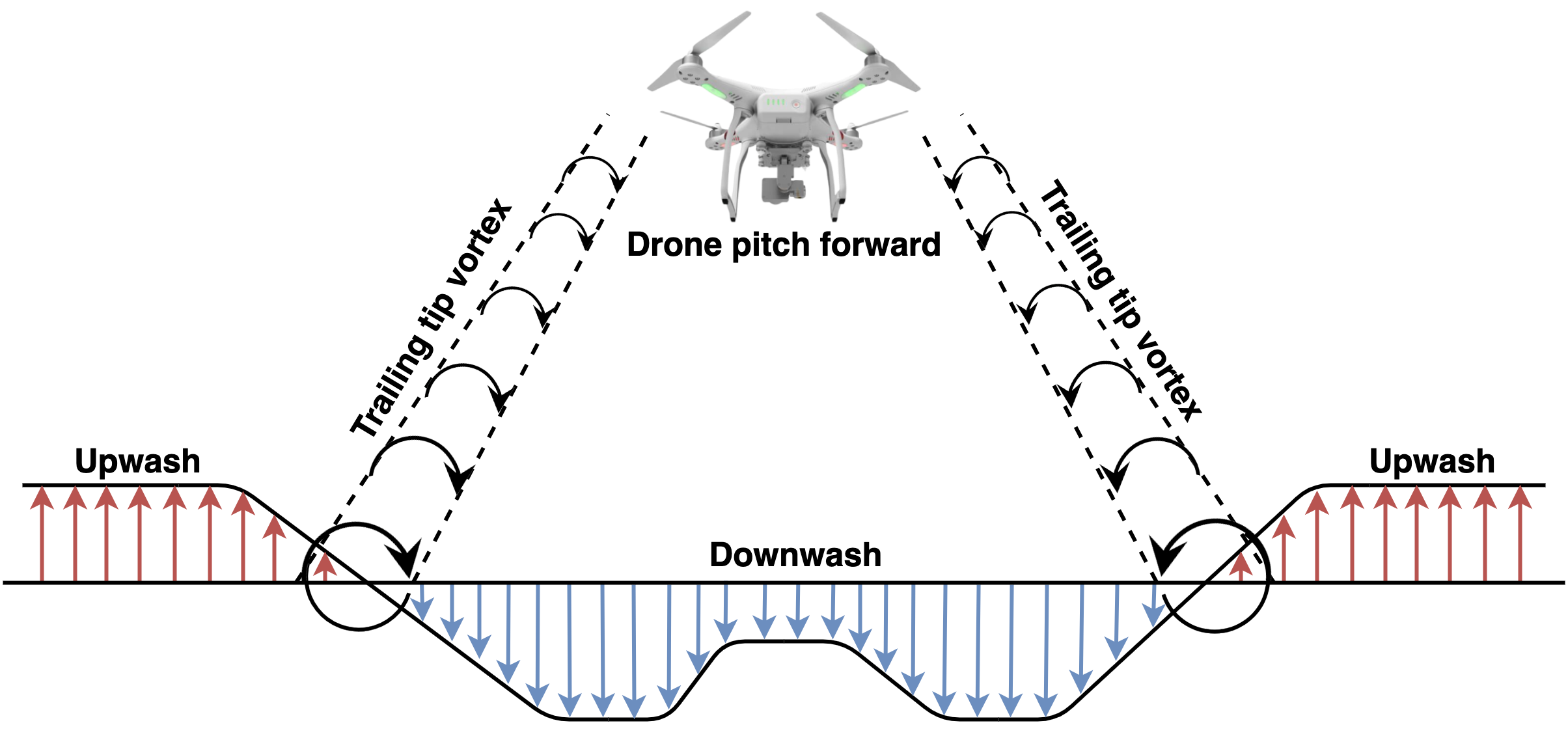}
\caption{View of upwash/downwash created by trailing vortices of the propellers}
\label{upwash}
\end{figure}

The study of trailing forces generated by wings have been largely studied in birds and fixed wing planes \cite{mirzaeinia2019energy}. As a bird flaps, a rotating vortex of air rolls off of each of its wingtips. This causes the air directly behind the bird to be pushed downwards (downwash) and the air off to the sides gets pushed upwards (upwash). If another bird flies in the upwash region, the bird will get a free lift, which allows it to save energy.  This causes the birds to fly in formations (like Vee) for this lift. A study showed that birds heart rates decreased when they fly in a Vee formation \cite{weimerskirch2001energy}. This demonstrates that the formation saves energy. This is specially important when the birds are migrating and travelling long distances.

In a similar manner, a quadcopter flying forward while maintaining its altitude also produces the same effect. As the quadcopter flies forward, the front propellers rotate slower than the back propellers, this allows the drone to pitch forward. This happens by decreasing the power in the front motors and increasing it in the back motors. The resulting net force will pitch the nose of the drone down and it will move forward. This variance in propellers rotating speeds along with the orientation produces upwash and downwash trailing forces (Fig \ref{upwash}). \looseness=-1

\begin{figure}
\centering
\includegraphics[width=0.7\linewidth]{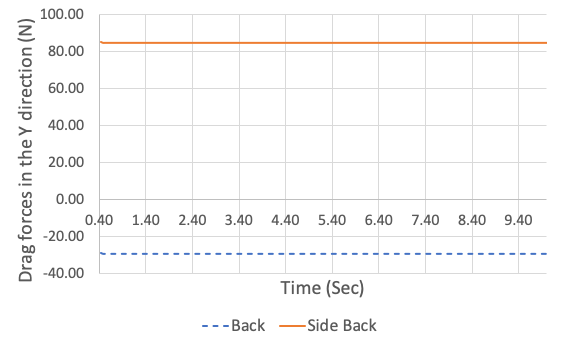}
\caption{Lift on Back and Side Back drones caused by Upwash/ Downwash Forces }
\label{upwashchart}
\end{figure}

We use Ansys Discovery Live to prove that these forces occur in real scenarios. We use a DJI Phantom 3 Model and create rotating walls on the propellers to study the behavior. We set the front propellers to rotate at a speed of 10 rev/s and the back propellers at a speed of 15.3 rev/s. We then test the drag forces in the y-axis, i.e lift forces, on another drone. First, we test when the drones are aligned infront of each other directly. Second, we test when the back drone is sliding a bit to the side of the front drone. The results as shown in Fig.\ref{upwashchart} show that a quadcopter in these settings act similar to the birds. The drone slightly to the side experienced a lift force in the positive direction (upwash). Whereas the drone directly at the back experienced a downward force in the negative direction (downwash). This behavior raises the need to fly the drones in formations.

\subsubsection{Drag Forces and Swarm Formations.}

\begin{figure}
\centering
\includegraphics[width=0.7\linewidth]{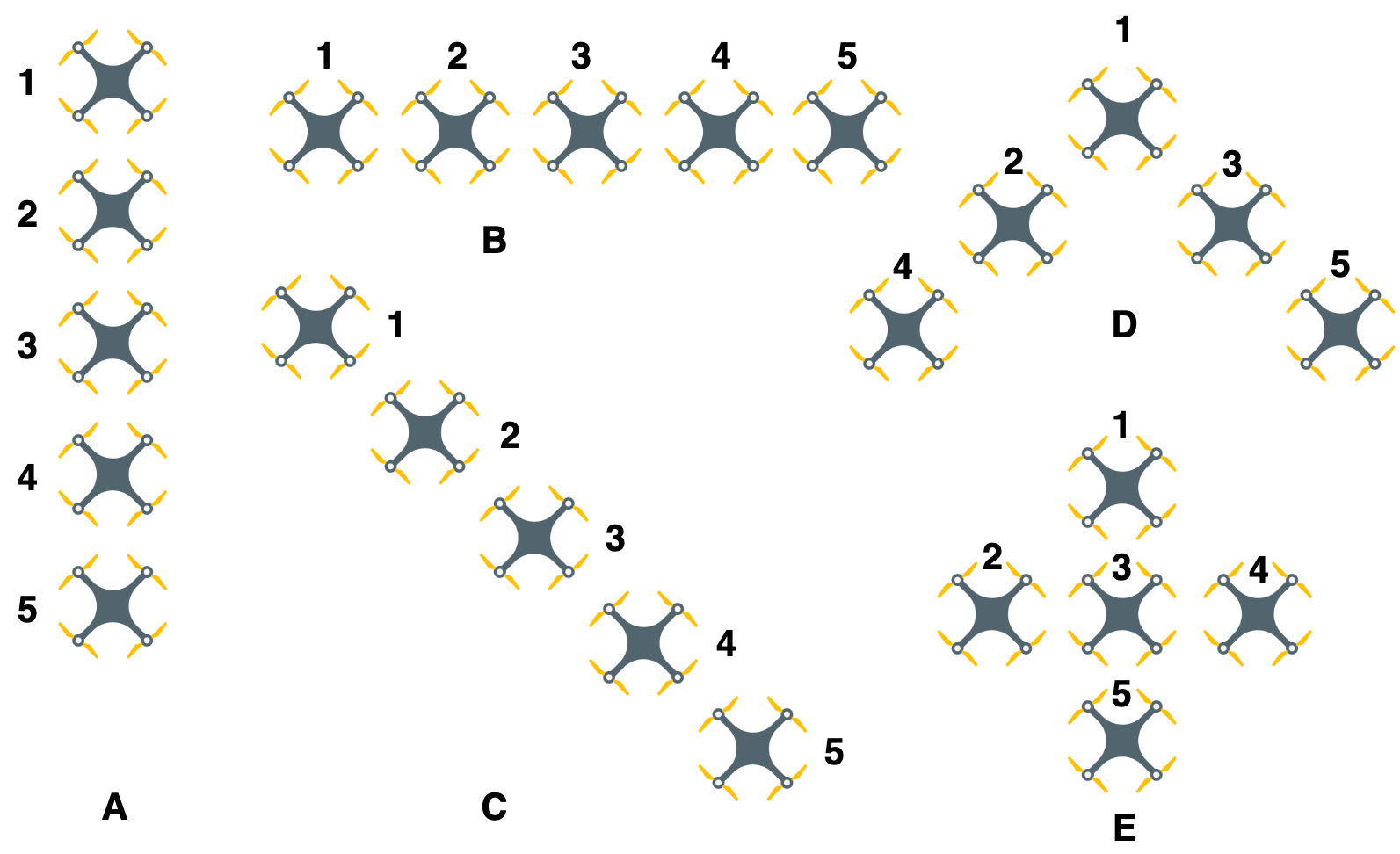}
\caption{Different flight formations for a swarm of drones. A: Column, B: Front, C: Echelon, D: Vee, E: Diamond}
\label{formations}
\end{figure}

There are different popular flight formations taken by planes in military and by migrating birds. We test how the formations affect the aerodynamic drag forces on the surfaces of the quadcopter. Then, we test the effect of different wind speeds on the drag forces on the swarm. Later, we test the drag forces on individual drones in the swarm and the overall average drag on the swarm. Finally, we compute the energy consumption of each formation and decide which formation is best under which wind condition. We test three wind directions Front, Right, and Left. We do not consider  winds coming from the back of the drone as they reduce the drag and in fact help the drone fly forward. In this paper, we study five different swarm formations shown in Fig.\ref{formations}. The formations are: Column, Front, Echelon, Vee, and Diamond. We assume a swarm is made up of five quadcopters as shown in Fig. \ref{formations}. The results are shown in the drag forces experiments subsection \ref{resultsdrag}. We use Ansys Discovery Live to carry out the expirements. A screenshot from the software is shown in Fig. \ref{ansysv}.\looseness=-1


Our goal is to drive the selection process from the formation of the swarm. As mentioned earlier, an $SDaaS$ service selection is the process of deciding whether a skyway segment delivers the functional requirement of an $SDaaS$ service. In other words, we want to select the skyway segments that have successful deliveries between the nodes based on the chosen formation. A successful delivery is a delivery where all the members of the swarm can reach the next node without consuming all of their battery capacity. An optimal selection is a selection where we maximize the reserved energy taken in a flight in a skyway segment. 

\subsection{SDaaS Selection Types}
There are two types of selections we propose in this paper: Fixed and Adaptive. We create two algorithms for each selection type. A selection consists of two main components. The first is the selection of the swarm formation. The second is the selection of the $SDaaS$ services. We assume the drones in the swarm are already selected. We also assume that all the drones in the swarm carry the same payload. We also assume that the drones fly in a relatively similar altitude throughout the journey. Hence, the effect of varying air pressure is minimal and is not considered in this work. We focus on studying the effect of formations on the service selection.\looseness=-1

\begin{enumerate}
    \item \textbf{Fixed SDaaS Selection}\\
    In a \emph{fixed selection} all wind conditions of all the services connecting the source and destination nodes are averaged. We take the average wind speed value between all skyway segments and the most common wind direction. We use this averaged value to choose the best fixed formation that will traverse the network without changing. Then, we check all the services, i.e. skyway segments, where \emph{all} the drones in the swarm will have successful flights to the next node. A successful flight is a flight where \emph{all} the drones in the swarm have sufficient power to reach the destination $\forall E_c<= 100\%$. The total energy consumption $E_c$ is equal to the sum of energy consumption due to flight range, wind drag forces, and upwash/downwash forces, $E_c= E_{fr}+E_{drag}+E_{upwash/downwash}$. Algorithm \ref{fixedAlg} describes the fixed selection process. \looseness=-1

    \item \textbf{Adaptive SDaaS Selection}\\
    In an \emph{adaptive selection}, the formation of the swarm can change at every skyway segment. In this type of selection, we look at every skyway segment and choose the best formation accordingly. The best formation is the formation that reduces the energy consumption due to wind drag force $E_{drag}$ and upwash/downwash forces $E_{upwash/downwash}$. Then, we check all the services, i.e. skyway segments, where \emph{all} the drones in the swarm will have successful flights to the next node. A successful flight is a flight where \emph{all} the drones in the swarm have sufficient power to reach the destination $\forall E_c<= 100\%$. Note here that selecting the optimal formation does not guarantee a successful flight. This is because the energy consumption is affected by other factors like the flight range and payload. The total energy consumption $E_c$ is equal to the sum of energy consumption due to flight range, wind drag forces, and upwash/downwash forces. $E_c= E_{fr}+E_{drag}+E_{upwash/downwash}$. Algorithm \ref{adaptiveAlg} describes the adaptive selection process.\looseness=-1
\end{enumerate}

\begin{center}

\begin{algorithm}
 \caption{Fixed Service Selection Algorithm}
 \label{fixedAlg}
 \begin{algorithmic}[1]
 \renewcommand{\algorithmicrequire}{\textbf{Input:}}
 \renewcommand{\algorithmicensure}{\textbf{Output:}}
 \REQUIRE $All_{SDaaS}$
 \ENSURE  $Sel_{SDaaS}$
 \STATE $Sel_{SDaaS}$ = [$\phi$]
    \STATE $TWindS$ = 0, $FrontWindDir$=0, $RightWindDir$=0, $LeftWindDir$=0
    \FOR{$SDaaS$ in $All_{SDaaS}$ }
        \STATE $TWindS$ += $SDaaS[WindS]$
        \IF{$SDaaS[WindDir]$ == Front}
            \STATE $FrontWindDir$ +=1
        \ELSIF{$SDaaS[WindDir]$ == Right}
            \STATE $RightWindDir$ +=1
        \ELSIF{$SDaaS[WindDir]$ == Left}
            \STATE $LeftWindDir$ +=1
        \ENDIF
    \ENDFOR
    \STATE $AvgWindS$= $TWindS$/Size($All_{SDaaS}$)
    \STATE $MaxWindDir$= type(Max($FrontWindDir$, $RightWindDir$, $LeftWindDir$))
    \STATE $SelectedFormation=$\textbf{SelectFormation}($AvgWindS$,$MaxWindDir$)
    \FOR{$SDaaS$ in $All_{SDaaS}$}
        \STATE $E_c= E_{fr}+E_{drag}+E_{upwash/downwash}$
        \IF{$\forall E_c$ $<$ 100}
            \STATE \textbf{append}($Sel_{SDaaS}$, $SDaaS$ )
        \ENDIF
    \ENDFOR
 \RETURN $Sel_{SDaaS}$
 \end{algorithmic}
 \end{algorithm}
     
\end{center}

\begin{algorithm}
 \caption{Adaptive Service Selection Algorithm}
 \label{adaptiveAlg}
 \begin{algorithmic}[1]
 \renewcommand{\algorithmicrequire}{\textbf{Input:}}
 \renewcommand{\algorithmicensure}{\textbf{Output:}}
 \REQUIRE $All_{SDaaS}$
 \ENSURE  $Sel_{SDaaS}$
 \STATE $Sel_{SDaaS}$ = [$\phi$]
    \FOR{$SDaaS$ in $All_{SDaaS}$ }
       \STATE $SelectedFormation$=\\\textbf{SelectFormation}($SDaaS[WindS]$,$SDaaS[WindDir]$)
       \STATE $E_c= E_{fr}+E_{drag}+E_{upwash/downwash}$
        \IF{$\forall E_c$ $<$ 100}
            \STATE \textbf{append}($Sel_{SDaaS}$, $SDaaS$ )
        \ENDIF
    \ENDFOR
 \RETURN $Sel_{SDaaS}$
 \end{algorithmic}
 \end{algorithm}

\section{Experimental Results and Discussion}
In this section, we discuss the results of drag forces on a swarm under different wind conditions. We conduct a set of experiments to evaluate the controlling attributes including wind speed and wind direction. We then conclude which formation is best under which condition and how much energy is consumed. In addition, we evaluate the performance of the proposed selection algorithms in terms of energy consumption and execution. 

\subsection{Drag and power consumption of different formations}
\label{resultsdrag}
Assuming that the swarm size is 5, i.e. the swarm composes of a set of five identical drones, we run Ansys software to test the drag forces on every individual drone in the swarm numbered in Fig. \ref{formations}. We used a DJI Phantom 3 model to run our tests\footnote{DJI Phantom 3 3D model: https://grabcad.com/library/dji-phantom3-1}. We then compute the average energy consumption of the full swarm. We fix the wind speed to Beaufort scale 5, i.e. 9.35m/s to test the drag from different wind directions. We also assume that the drones are flying at a speed of 15.6m/s. Table \ref{tab:drag} summarizes the results of the first experiment. We can notice that the best formation that reduces the drag forces when the wind comes from the front is the \emph{Column} formation. On the other hand, the \emph{Front} formation performed best when the wind is coming from the right or left. Note that we do not consider the upwash/downwash forces in this experiment. If we look at the drag forces of individual drones, we can get insights on which drones face most drag due to wind: 

\begin{itemize}
    \item In a \textbf{Column Formation}, the front drones protect the drones at the back from the front wind. Hence, the drag is in a decreasing manner from drone 1 to 5. However, when the wind comes from the right or left, the drones at the edge (1 and 5) face more drag than the inner drones. This is because we compute the drag on all the faces of the drone. The side faces of the inner drones are protected from the wind by the neighbouring drones.
    \item In a \textbf{Front Formation}, a reciprocal behaviour to the column formation is seen. In a front wind, the edge drones (1 and 5) face more drag than the inner drones. This is because the side faces of the inner drones are protected from the wind by the neighbouring drones. In a right and left winds, the drag is in a decreasing manner with the maximum drag at the first drone facing the wind direction. The rest of the drones are protected by the drones ahead.
    \item The \textbf{Echelon Formation} is one of the highest formations facing drag forces. However, we can notice, that the drones consume energy in a decreasing manner from drones 1 to 5 in a front and left wind direction as some parts of the drones are covered by the front drones. The opposite behavior  is noted for the right wind direction with drone 5 first facing the wind.
    \item In a \textbf{Vee Formation}, the drones facing the wind experience more drag. With a front wind, drone 1 experiences maximum drag followed by drones 4 and 5 at the back. This aligns with the study made in \cite{mirzaeinia2019energy} which found that fixed-wing drones in a Vee formation experience more drag at the front and back of the swarm compared to the inner drones. When wind comes from the right or the left, the drones closer facing the wind experience more drag compared to the rest of the swarm which are protected in the swarm.
    \item In a \textbf{Diamond Formation}, the drag forces on drones 3 and 5 are significantly lower than the others in the front wind. \emph{This can be important in applications where the most valuable delivery items are carried by these specific drones that fly in the formation to be protected from the wind}. Similar behavior is noticed in drones 2 and 3 with the right wind and drones 3 and 4 with the left wind. 
\end{itemize}

In the second experiment, we evaluate the \emph{combined effect of drag forces with the upwash/downwash forces}. We can see from Table \ref{tab:upawshDragEnergy} that the formations that consume the least amount of energy is the \emph{Vee} formation for the front wind and the  \emph{Diamond} formation for right and left winds. Although the \emph{Column and Front} formation performed better when only the drag forces are computed, the \emph{Vee and Diamond} outperformed them due to the upwash/downwash forces from the drones swarm positions. The energy consumption is computed using Equation \ref{equation}. \looseness=-1

\begin{figure}[ht]
\centering
\includegraphics[width=0.8\linewidth]{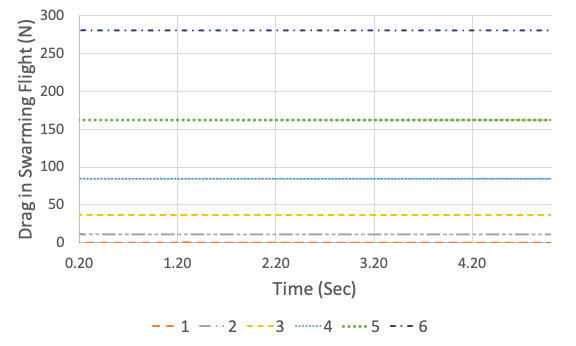}
\caption{Drag Forces on Vee formation for Different Beaufort Numbers }
\label{Beaufort}
\end{figure}

In the third experiment, we evaluate the effect of varying different wind speeds according to Beaufort scale. The results of this experiment on the Vee formation are shown in Fig.\ref{Beaufort}. As shown, Beaufort scales 0 and 1 have minimal effect on the drag value. As the scale increases the drag value approximately doubles at each scale number. Similar behavior is seen for the different formations. Since the behavior is similar in all formations, the selection of the formation mainly depends on the wind direction rather than the speed. The drag values based on the speed are used for the service selection in the next subsection.\\

\begin{table}
\centering
\small
\begin{tabular}{|l|l|l|l|l|}
\hline
\multicolumn{2}{|l|}{\begin{tabular}[c]{@{}l@{}}\textbf{Swarm Formation/} \\ \textbf{Wind Direction}\end{tabular}} & \textbf{Front}                & \textbf{Right}                & \textbf{Left}                 \\ \hline
\multirow{6}{*}{\textbf{Column}}                        & D1 Drag                                & 134.40                        & 187.38                        & 188.68                        \\ \cline{2-5} 
                                                        & D2 Drag                                & 93.88                         & 171.58                        & 169.28                        \\ \cline{2-5} 
                                                        & D3 Drag                                & 90.39                         & 167.57                        & 168.24                        \\ \cline{2-5} 
                                                        & D4 Drag                                & 81.40                         & 184.36                        & 184.45                        \\ \cline{2-5} 
                                                        & D5 Drag                                & 79.29                         & 209.68                        & 210.54                        \\ \cline{2-5} 
                                                        & \textbf{Avg Drag}                      & {\ul \textit{\textbf{95.87}}} & \textbf{184.11}               & \textbf{184.24}               \\ \hline
\multirow{6}{*}{\textbf{Front}}                         & D1 Drag                                & 187.40                        & 79.93                        & 134.22                        \\ \cline{2-5} 
                                                        & D2 Drag                                & 171.58                        & 82.22                          & 94.26                         \\ \cline{2-5} 
                                                        & D3 Drag                                & 166.77                        & 90.55                         & 92.45                         \\ \cline{2-5} 
                                                        & D4 Drag                                & 183.86                        &  93.83                         &  80.83                         \\ \cline{2-5} 
                                                        & D5 Drag                                & 210.23                        &  133.78                        &  79.21                         \\ \cline{2-5} 
                                                        & \textbf{Avg Drag}                      & \textbf{183.97}               & {\ul \textit{\textbf{96.06}}} & {\ul \textit{\textbf{96.19}}} \\ \hline
\multirow{6}{*}{\textbf{Echelon}}                       & D1 Drag                                & 178.64                        & 135.12                         & 179.64                        \\ \cline{2-5} 
                                                        & D2 Drag                                & 187.39                        & 165.74                        & 187.45                        \\ \cline{2-5} 
                                                        & D3 Drag                                & 178.07                        & 179.10                        & 178.07                        \\ \cline{2-5} 
                                                        & D4 Drag                                & 165.74                        &  187.39                        & 165.74                        \\ \cline{2-5} 
                                                        & D5 Drag                                & 136.22                        & 179.64                        & 135.94                        \\ \cline{2-5} 
                                                        & \textbf{Avg Drag}                      & \textbf{169.21}               & \textbf{169.40}               & \textbf{169.37}               \\ \hline
\multirow{6}{*}{\textbf{Vee}}                           & D1 Drag                                & 184.41                        & 379.84                        & 378.40                        \\ \cline{2-5} 
                                                        & D2 Drag                                & 133.87                        & 115.35                        & 332.57                        \\ \cline{2-5} 
                                                        & D3 Drag                                & 131.27                        & 330.57                        & 114.45                        \\ \cline{2-5} 
                                                        & D4 Drag                                & 152.18                        & 160.51                        & 354.83                        \\ \cline{2-5} 
                                                        & D5 Drag                                & 167.81                        & 354.92                        & 164.19                         \\ \cline{2-5} 
                                                        & \textbf{Avg Drag}                      & \textbf{153.91}               & \textbf{268.24}               & \textbf{268.89}               \\ \hline
\multirow{6}{*}{\textbf{Diamond}}                       & D1 Drag                                & 180.90                        & 180.52                        & 182.24                        \\ \cline{2-5} 
                                                        & D2 Drag                                & 196.50                        & 68.65                        & 193.00                        \\ \cline{2-5} 
                                                        & D3 Drag                                & 91.67                         & 91.09                         & 92.13                         \\ \cline{2-5} 
                                                        & D4 Drag                                & 182.85                        & 184.65                        & 67.56                        \\ \cline{2-5} 
                                                        & D5 Drag                                & 68.98                         & 195.30                         & 181.85                         \\ \cline{2-5} 
                                                        & \textbf{Avg Drag}                      & \textbf{144.18}               & \textbf{144.04}               & \textbf{143.36}               \\ \hline
\end{tabular}%
\caption{Individual and Average Drag Values for Swarm Formations}
\label{tab:drag}
\end{table}

\begin{table}
\centering
\begin{tabular}{|l|l|l|l|}
\hline
                 & \textbf{Front}         & \textbf{Right}         & \textbf{Left}          \\ \hline
\textbf{Column}  & 1858.78                & 3235.33                & 3237.27                \\ \hline
\textbf{Front}   & 2869.90                & 1498.56                & 1500.64                \\ \hline
\textbf{Echelon} & 1576.44                & 1579.35                & 1578.85                \\ \hline
\textbf{Vee}     & {\ul \textbf{1337.69}} & 3121.24                & 3131.37                \\ \hline
\textbf{Diamond} & 1367.50                & {\ul \textbf{1365.35}} & {\ul \textbf{1354.63}} \\ \hline
\end{tabular}%
\caption{Average Energy Consumption due to Drag Forces and Upwash/Downwash Forces in Watts}
\label{tab:upawshDragEnergy}
\end{table}

\subsection{Selection algorithms evaluation}
The dataset used in the experiments is an urban road network dataset from the city of London with nodes representing intersections and edge length representing the distances between the nodes \cite{karduni2016protocol}. For the experiments, we took a sub-network of connected nodes with size 2732 to mimic a possible arrangement of a skyway network. We then synthesize different wind speeds and wind directions for every segment. We assume that the drones model is DJI Phantom 3 flying at a speed of 15.6 m/s. The rest of the experimental variables are summarized in Table \ref{tab:variables}. These variables are used to compute the energy consumption based on the flight range, drag forces, and upwash/downwash forces.\looseness=-1

\begin{table}[ht]
\caption{Experimental Variables}
\centering
\label{tab:variables}
\resizebox{0.5\linewidth}{!}{%
\begin{tabular}{l|l}
\hline
Variable         & Value         \\ \hline
Network Size     & 2732 nodes    \\
Drones Type      & DJI Phantom 3 \\
Battery Capacity & 4480 mAh      \\
Voltage          & 15.2 V        \\
Drone Speed      & 15.6 m/s      \\ \hline
\end{tabular}%
}
\end{table}

\begin{figure}[ht]
\centering
\includegraphics[width=0.8\linewidth]{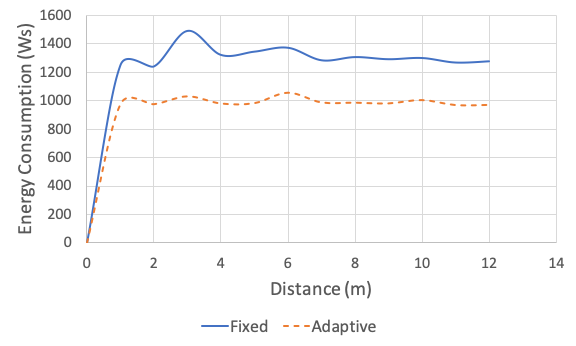}
\caption{Energy Consumption at Different Distances}
\label{energyDistanceFig}
\end{figure}

In the first experiment, we evaluate the energy consumption of both selection algorithms, i.e. Fixed and Adaptive. We group the output of the algorithms, i.e. the selected services, based on their distances. Unselected services are discarded as the swarm will fail its journey due to lack of battery power in these services. The x-axis in Fig.\ref{energyDistanceFig} represents the distance and the y-axis represents the energy consumption due to drag forces and upwash/downwash forces. As shown in the graph, the proposed adaptive algorithm outperforms the fixed algorithm significantly. The adaptive algorithm saves more energy since the optimal formation that saves the most energy is selected at each skyway segment. \looseness=-1

\begin{figure}[ht]
\centering
\includegraphics[width=0.8\linewidth]{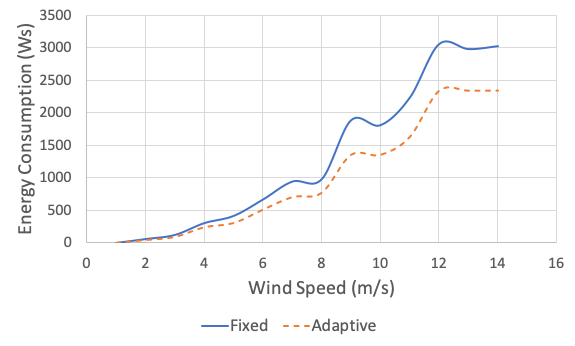}
\caption{Energy Consumption at Different Wind Speeds}
\label{energyWindFig}
\end{figure}

In the second experiment, we repeat the first experiment but we group the services based on their wind speeds rather than distance. This will allow us to see the performance of both algorithms under different wind conditions. As shown in Fig. \ref{energyWindFig} the adaptive algorithm outperforms the fixed algorithm in terms of energy consumption. This is because the drones in the fixed algorithm select the formation based on average wind conditions which may not be the optimal solution under different wind conditions. \looseness=-1

\begin{figure}[ht]
\centering
\includegraphics[width=0.8\linewidth]{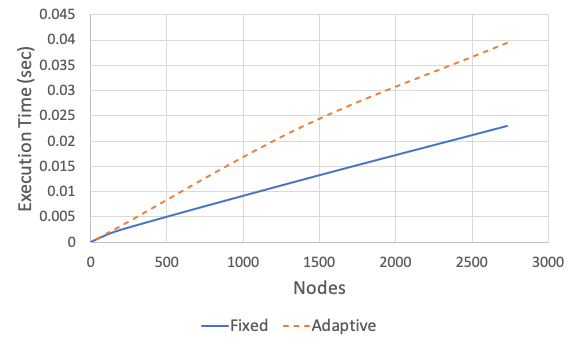}
\caption{Fixed Vs Adaptive Algorithms Execution Times}
\label{executionFig}
\end{figure}

Although the adaptive algorithm outperformed the fixed algorithm, it comes with the cost of execution time. This is shown in Fig.\ref{executionFig} where the x-axis represents the number of nodes of different sub-networks of our full network. As the network size increases the adaptive algorithm takes more execution time compared to the fixed algorithm. We conclude, that depending on the application, the choice of algorithm implementation would be different. For example, in the case of an emergency where we need to ensure a fast delivery with the least number of stops for recharging, more resources may be utilized for the computation. The computations are usually done on edge nodes. On the other hand, if the delivery time is flexible but the number of resources for computation is limited, then the fixed algorithm would be the better choice.\looseness=-1

\section{Conclusion}

We proposed a Swarm-based Drone-as-a-Service selection model for delivery services. We identified two types of selections namely: Fixed and Adaptive. Two algorithms are proposed for each type. The two algorithms take extrinsic constraints, explicitly the wind conditions, into consideration to select the optimal formation and services.  We studied the effect of different swarm formations on the drag forces and upwash/downash forces caused by the wind and the drones arrangement. Experimental results show that the Adaptive algorithm outperformed the Fixed algorithm in terms of energy consumption. In the future work, we will build upon and extend this work to compose the best set of selected services in terms of energy consumption and delivery time from a source node to a destination node.

\bibliographystyle{ACM-Reference-Format}
\bibliography{Scholar}

\end{document}